\documentclass[paper]{ieice}
\usepackage[fleqn]{amsmath}
\usepackage{newtxtext}
\usepackage[varg]{newtxmath}
\usepackage{cite}
\usepackage{mathtools}

\field{D}
\title{Information Geometric Self-Organization at the Edge of
  Stability in High-Capacity Kernel Associative Memories}

\authorlist{%
  \authorentry{Akira Tamamori}{m}{AIT}\MembershipNumber{1512145}
  \affiliate[AIT]{The author is with the Faculty of Information
    Science, Aichi Institute of Technology, Aichi 470-0392, Japan.}
}

\received{2015}{1}{1}
\revised{2015}{1}{1}

\begin{document}
\maketitle

\begin{summary}
  High-capacity associative memories based on Kernel Logistic
  Regression (KLR) exhibit exceptional storage capabilities and
  robustness. Previous empirical studies identified a hyperparameter
  regime, the ``Ridge of Optimization,'' where attractor stability is
  maximized. However, the geometric nature of this regime and the
  optimization dynamics required to reach it have remained unclear. In
  this paper, we investigate the static geometry of the parameter
  space and the learning trajectory of Gradient Descent (GD) in
  KLR-trained Hopfield networks. Using the eigenvalue spectrum of the
  Hessian, we reveal that the Ridge corresponds to a phase boundary
  located adjacent to a rank-1 spectral collapse, acting as a
  geometric singularity where the principal curvature is massively
  amplified. Furthermore, we demonstrate that the learning dynamics
  exhibit a transient self-stabilizing behavior driven by the Edge of
  Stability (EoS) phenomenon. Rather than seeking flat regions, the
  network parameters are driven toward a state where the local
  curvature dynamically equilibrates near the stability limit dictated
  by the learning rate, allowing the optimization to survive the
  initial instability. We provide analytical derivations for both the
  rank-1 asymptotic collapse and the dynamic feedback loop governing
  this equilibration. These findings suggest that optimal,
  high-capacity memory representations are not formed in flat minima,
  but are dynamically sculpted at the highly curved boundaries of
  geometric singularities.
\end{summary}

\section{Introduction}
Associative memory models, conventionally typified by the Hopfield
network~\cite{Hopfield1982}, function by storing data patterns as
stable fixed points within a defined energy landscape. While the
classical model using Hebbian learning provides a foundational
framework, its practical application is restricted by a storage
capacity limit of $P \approx 0.14N$~\cite{Amit1985}.

Recent research has explored methods to overcome this limitation. One
approach, known as Modern Hopfield Networks (MHNs), increases capacity
by modifying the energy function to include polynomial or exponential
interaction terms \cite{Krotov2016, Ramsauer2021}. An alternative
approach relies on discriminative learning algorithms while
maintaining a standard quadratic energy structure in a feature
space. Our previous works demonstrated that Hopfield networks trained
via Kernel Logistic Regression (KLR) achieve high storage capacities
exceeding classical limits~\cite{tamamori_letter,
  tamamori_nolta}. Furthermore, a structural regime termed the ``Ridge
of Optimization'' was identified, where the network localizes its
parameter distribution to maintain stable attractors under high memory
load conditions~\cite{tamamori_nolta_b}.

Despite the empirical validation of this high-capacity regime, the
underlying optimization dynamics remain unclear. Specifically, it is
not well understood how a standard optimization algorithm, such as
Gradient Descent (GD), navigates the high-dimensional parameter space
to find this specific ridge structure. In the context of deep
learning, the geometry of the loss landscape and its interaction with
learning dynamics are often analyzed using the Hessian or the Fisher
Information Matrix (FIM)~\cite{Amari2016}. Recent studies on the
learning dynamics of neural networks have highlighted phenomena such
as the Edge of Stability (EoS)~\cite{Cohen2021, Damian2023}, where the
sharpness of the loss landscape increases until it reaches the
stability limit of GD, leading to oscillatory behavior. However, the
geometric relationship between the emergence of the Ridge in kernel
associative memories and such optimization dynamics has not been
systematically investigated.

This paper investigates the learning dynamics of KLR-trained Hopfield
networks from the perspectives of information geometry and dynamical
systems. We aim to clarify the geometric nature of the Ridge of
Optimization and the mechanism by which GD converges to this
state. The main contributions of this paper are as follows:

\begin{enumerate}
\item We geometrically characterize the Ridge of Optimization using
  the eigenvalue spectrum of the Hessian. By overlaying the memory
  retrieval performance on the geometric phase diagram, we demonstrate
  that the high-capacity regime is located adjacent to a rank-1
  spectral collapse region. This indicates that optimal memory
  formation occurs near a singularity where the principal curvature is
  highly amplified.
  
\item We show that the learning trajectory of GD exhibits a
  self-organizing behavior driven by the EoS phenomenon. Through
  theoretical modeling and empirical analysis of learning rate
  variations, we verify that the network parameters are driven toward
  a state where the local curvature (the maximum eigenvalue of the
  Hessian) dynamically equilibrates near the stability limit dictated
  by the learning rate.
  
\end{enumerate}

The remainder of this paper is organized as follows. Section 2 reviews
the network model, the information-geometric formulation, and the
experimental setup. Section 3 analyzes the static geometry of the
parameter space using the Hessian spectrum and its relation to
retrieval performance. Section 4 presents empirical results on the
learning dynamics of GD, focusing on the self-stabilization
mechanism. Section 5 provides theoretical analyses of the spectral
collapse and the dynamic equilibration near the stability
limit. Section 6 discusses the implications for the flat minima
hypothesis and the limitations of our approach, and Section 7
concludes the paper.

\section{Background and Preliminaries}
This section briefly reviews the Kernel Logistic Regression (KLR)
Hopfield network model and introduces the geometric quantities used
for analyzing the learning dynamics.

\subsection{Kernel Logistic Regression Hopfield Network}
We consider an auto-associative memory network consisting of $N$
bipolar neurons, with the state vector denoted by
$\mathbf{s} \in \{-1, 1\}^N$. The network stores $P$ patterns, defined
as a set $\{\boldsymbol{\xi}^\mu\}_{\mu=1}^P$. The retrieval dynamics
are governed by the input potential $h_i(\mathbf{s})$ for each neuron
$i$, defined using a kernel function $K(\cdot, \cdot)$:
\begin{equation}
h_i(\mathbf{s}) = \sum_{\mu=1}^P \alpha_{\mu i} K(\mathbf{s}, \boldsymbol{\xi}^\mu),
\end{equation}
where $\boldsymbol{\alpha} = (\alpha_{\mu i})$ represents the dual
variable matrix. In this study, we employ the Gaussian Radial Basis
Function (RBF) kernel,
$K(\mathbf{x}, \mathbf{y}) = \exp(-\gamma\|\mathbf{x} -
\mathbf{y}\|^2)$, where $\gamma$ is a scalar parameter controlling the
locality of the kernel mapping.

The dual variables for each neuron $i$, denoted by the vector
$\boldsymbol{\alpha}_i = [\alpha_{1i}, \dots, \alpha_{Pi}]^\top$, are
determined through supervised learning. The learning objective is to
predict the target state $y_\mu = (\xi^\mu_i + 1)/2 \in \{0, 1\}$ from
the input pattern $\boldsymbol{\xi}^\mu$. This is formulated as the
minimization of an $L_2$-regularized negative log-likelihood function
$L(\boldsymbol{\alpha}_i)$:
\begin{align}
  L(\boldsymbol{\alpha}_i) &= L_{\text{logistic}}(\boldsymbol{\alpha}_i) + \frac{\lambda}{2} \boldsymbol{\alpha}_i^\top \mathbf{K} \boldsymbol{\alpha}_i,\\
  L_{\text{logistic}}(\boldsymbol{\alpha}_i)
  &= -\sum_{\mu=1}^P \left[ y_\mu \log(\sigma(h_i(\boldsymbol{\xi}^\mu))) \right. \nonumber \\
  &\quad\quad\quad\quad \left.+ (1 - y_\mu) \log(1 - \sigma(h_i(\boldsymbol{\xi}^\mu))) \right],
\label{eq:objective}
\end{align}
where $\sigma(z) = 1/(1 + e^{-z})$ is the logistic sigmoid function,
$\mathbf{K}$ is the $P \times P$ kernel Gram matrix with entries
$K_{\mu\nu} = K(\boldsymbol{\xi}^\mu, \boldsymbol{\xi}^\nu)$, and
$\lambda$ is the weight decay parameter. This optimization is
typically performed using iterative methods such as Gradient Descent
(GD). In the following analysis, we focus on the parameter trajectory
$\boldsymbol{\alpha}_i$ for a single representative neuron and omit
the subscript $i$ for notational simplicity.

Previous studies on KLR-trained Hopfield networks identified a
specific structural regime, termed the ``Ridge of
Optimization''~\cite{tamamori_nolta_b}. This ridge represents an
optimal operating region in the hyperparameter space (spanning the
storage load $P/N$ and the kernel locality $\gamma$) where the local
stability of the attractors is maximized, ensuring high storage
capacity and robust error correction. In this paper, we focus our
analysis on this representative regime to elucidate the geometric and
dynamical mechanisms underlying such high performance.

\subsection{Fisher Information Matrix and the Hessian}
To analyze the geometric properties of the parameter space and the
stability of the optimization dynamics, we distinguish between two
related matrices: the Fisher Information Matrix (FIM) and the Hessian
of the objective function.

From an information-geometric perspective~\cite{Amari2016}, the
unregularized logistic model defines a statistical manifold. The
intrinsic geometry of this manifold is characterized by the FIM,
denoted by $\mathbf{G}(\boldsymbol{\alpha})$, which is equivalent to
the Hessian of the negative log-likelihood component
$L_{\text{logistic}}(\boldsymbol{\alpha})$. For the KLR model, the FIM is
given by:
\begin{equation}
  \mathbf{G}(\boldsymbol{\alpha}) = \mathbf{K} \mathbf{D}(\boldsymbol{\alpha}) \mathbf{K},
\label{eq:fim}
\end{equation}
where $\mathbf{D}(\boldsymbol{\alpha})$ is a diagonal matrix
containing the prediction variances, with diagonal entries
$D_{\mu\mu} = p_\mu(1 - p_\mu)$ and output probability
$p_\mu = \sigma(h(\boldsymbol{\xi}^\mu))$. We utilize
$\mathbf{G}(\boldsymbol{\alpha})$ to analyze the spectral
concentration and the intrinsic dimensionality of the learned
representations.

On the other hand, the actual dynamics of GD are governed by the
curvature of the complete regularized objective function
$L(\boldsymbol{\alpha})$. The Hessian matrix of
$L(\boldsymbol{\alpha})$, denoted by
$\mathbf{H}(\boldsymbol{\alpha})$, is given by:
\begin{equation}
  \mathbf{H}(\boldsymbol{\alpha}) = \mathbf{G}(\boldsymbol{\alpha}) + \lambda \mathbf{K} = \mathbf{K} \mathbf{D}(\boldsymbol{\alpha}) \mathbf{K} + \lambda \mathbf{K}.
\label{eq:hessian}
\end{equation}

For stability analysis, the maximum eigenvalue of the Hessian,
$\lambda_{max}(\mathbf{H})$, dictates the local sharpness of the loss
landscape. In classical optimization theory, linear stability of GD
strictly requires $\lambda_{max}(\mathbf{H}) < 2/\eta$, where $\eta$
is the learning rate~\cite{LeCun1998}. Recently, it has been observed
that neural network training often drives the principal curvature
precisely to this theoretical threshold, leading to non-monotonic
optimization dynamics, a phenomenon termed the Edge of Stability
(EoS)~\cite{Cohen2021}.  In our experiments, because the
regularization parameter $\lambda$ is typically small (e.g.,
$\lambda=0.01$), the contribution of $\lambda \mathbf{K}$ is
relatively minor, and
$\mathbf{H}(\boldsymbol{\alpha}) \approx
\mathbf{G}(\boldsymbol{\alpha})$ holds in most functional
regimes. Nevertheless, to ensure theoretical rigor, we explicitly use
$\mathbf{H}(\boldsymbol{\alpha})$ when evaluating the stability limits
of the learning dynamics.

\subsection{Experimental Setup}
To systematically evaluate the geometric and dynamical properties of
the network, we conducted numerical simulations using random binary
patterns. Each element of the stored patterns $\boldsymbol{\xi}^\mu$
was independently drawn from $\{-1, 1\}$ with equal probability. The
patterns were generated using a fixed random seed (e.g., NumPy seed
42) to ensure reproducibility across different hyperparameter
settings. Unless otherwise specified, the network size was set to
$N = 100$ or $N=50$ depending on the computational requirements of the
specific analysis.

\begin{figure*}[t]
\begin{center}
\includegraphics[width=\linewidth]{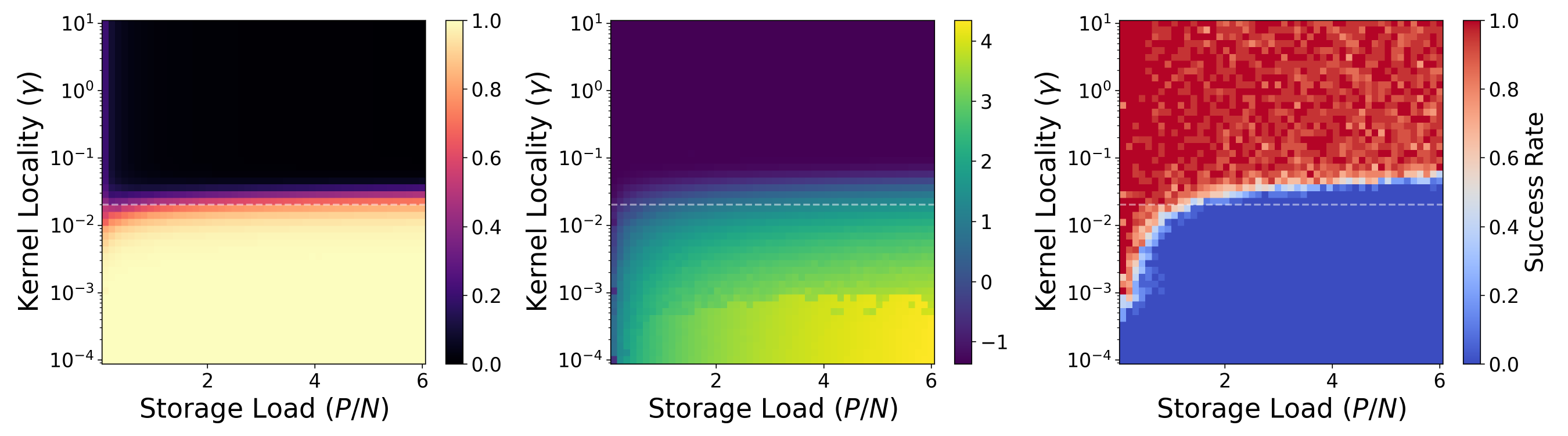}
\caption{Phase diagrams comparing the geometry of the statistical
  manifold and retrieval performance across storage loads ($P/N$) and
  kernel localities ($\gamma$). Parameters were optimized using
  L-BFGS-B (max 200 iterations, \texttt{ftol=1e-5}). Data points
  represent measurements from a single representative random pattern
  realization at each grid point. \textbf{(Left)} Spectral
  Concentration, $\lambda_{\max} /
  \mathrm{Tr}(\mathbf{H})$. \textbf{(Center)} Maximum Hessian
  Eigenvalue (log scale). \textbf{(Right)} Retrieval Success Rate from
  initial states with 20\% noise, evaluated over a subset of up to 10
  patterns per grid point.}
\label{fig:phase_diagram}
\end{center}
\end{figure*}

\textbf{Optimization Methods and Stopping Criteria:} For the dynamic
trajectory analysis (Section 4), we used full-batch GD. The initial
weights were initialized to zero
($\boldsymbol{\alpha}_0 = \boldsymbol{0}$). The learning rate was
fixed at $\eta = 0.1$ (unless varied for specific experiments), and
the $L_2$-regularization parameter was set to $\lambda = 0.01$. The
maximum number of training epochs was set to $300$. Preliminary
experiments confirmed that within 300 epochs, the macroscopic
oscillations characteristic of the Edge of Stability phase subside,
and the loss smoothly descends to a stabilized value, making it a
sufficient duration to observe the transition between the distinct
dynamical phases.

For the static geometric analysis requiring convergence to the exact
local minimum (Section 3), we utilized the L-BFGS-B
algorithm~\cite{Byrd1995}. The optimization was run with a maximum of
200 iterations and a strict tolerance for the objective function
change (\verb+ftol = 1e-5+).

\textbf{Phase Diagram Generation and Retrieval Evaluation:} The phase
diagrams (Section 3) were generated over a $30 \times 30$ grid of
storage loads ($P/N \in [0.1, 6.0]$) and kernel locality parameters
($\gamma \in [10^{-4}, 10^1]$) for a network of $N=50$. The plotted
values for spectral concentration and maximum curvature represent the
exact measurements from a single, representative random realization of
the pattern set at each grid point.

To evaluate the retrieval success rate at each grid point, we tested a
subset of 10 patterns (or $P$ patterns if $P < 10$). For each test
pattern, we generated a noisy initial state by randomly flipping bits
with an initial similarity of $m_0 = 0.8$ (i.e., 20\% noise). The
synchronous retrieval dynamics were iterated for a maximum of 30
steps. A recall was classified as successful only if the network state
perfectly converged to the original target pattern within these
steps. The reported success rate is the average over these tested
patterns.

\section{Phenomenology: The Geometry of the Ridge}
Previous empirical studies demonstrated that KLR Hopfield networks
achieve optimal storage capacity and noise robustness within a
specific hyperparameter regime, referred to as the Ridge of
Optimization~\cite{tamamori_nolta}. To establish a direct relationship
between this high-performance regime and the intrinsic geometry of the
statistical manifold, we systematically evaluated both the spectral
properties of the Hessian and the actual retrieval performance across
the hyperparameter space.

Based on the converged parameters $\boldsymbol{\alpha}^*$ obtained via
L-BFGS-B, we utilized two spectral metrics derived from the Hessian
$\mathbf{H}(\boldsymbol{\alpha}^*)$ to characterize the geometry,
alongside a direct measure of memory functionality:
\begin{enumerate}
\item \textbf{Spectral Concentration:} Defined as the ratio of the
  maximum eigenvalue to the trace of the Hessian,
  $\lambda_{\max}(\mathbf{H}) / \mathrm{Tr}(\mathbf{H})$. A value
  approaching $1.0$ indicates that the Hessian effectively collapses
  to a rank-1 matrix, implying extreme spectral anisotropy where the
  parameter space loses its multi-dimensional structure.
\item \textbf{Maximum Curvature:} Quantified by the logarithm of the
  maximum eigenvalue, $\log_{10} \lambda_{\max}(\mathbf{H})$. This
  value represents the steepness of the local loss landscape.
\item \textbf{Retrieval Success Rate:} The proportion of target
  patterns successfully recalled from initial states corrupted with
  20\% random bit-flip noise.
\end{enumerate}

Figure \ref{fig:phase_diagram} presents the resulting phase diagrams
for these three metrics. The phase diagrams reveal a striking
correspondence between the static geometry of the parameter space and
the dynamic memory performance. The right panel shows a clear phase
boundary separating the functional memory regime (red, success rate
$\approx 1.0$) from the overloaded regime where memory collapses
(blue, success rate $\approx 0.0$).

By comparing this performance boundary with the geometric metrics, we
observe that the memory collapse directly coincides with the geometric
collapse. In the lower-right region (high $P/N$ and low $\gamma$), the
left panel shows that the spectral concentration reaches $1.0$ (white
region), indicating that the Hessian becomes a rank-1 matrix. In this
state, the statistical manifold loses the multi-dimensional structure
necessary for separating distinct patterns. Concurrently, the center
panel shows that the maximum curvature $\lambda_{\max}$ increases
exponentially toward this region.

Importantly, the previously identified Ridge of Optimization
(indicated by the dashed line at $\gamma \approx 0.02$) serves as a
representative operating point that balances stability and
capacity. As the storage load increases, this line closely approaches
the boundary of the rank-1 collapse region.  Note that at extremely
high loads ($P/N \gtrsim 4.0$), the performance boundary slightly
shifts upward toward higher $\gamma$ values. This is a theoretically
expected behavior, as greater memory congestion necessitates stronger
kernel locality to mitigate inter-pattern interference. Consequently,
the representative line at $\gamma = 0.02$ eventually intersects the
memory collapse regime at these extreme loads.  At this boundary, the
network maintains the minimal spectral diversity required for pattern
separation, while the principal curvature $\lambda_{\max}$ is highly
amplified (e.g., $\lambda_{\max} > 100$).

These results provide direct evidence that the high-capacity memory
regime is situated adjacent to a region of extreme spectral
degeneracy. Optimal memory performance is achieved not in a
geometrically flat region, but at the edge of a structural transition,
where the learning algorithm maximizes the principal restorative force
just before the pattern separation capability is lost.

\section{Learning Dynamics: Transient Self-Stabilization}
The static geometric analysis in Section 3 indicates that
high-capacity memory representations are formed near a boundary
characterized by extreme spectral degeneracy and high curvature. To
investigate how a standard first-order optimization method navigates
this steep landscape, we analyzed the learning trajectory of GD.  As
introduced in Section 2, optimization stability is fundamentally
governed by the maximum eigenvalue of the Hessian,
$\lambda_{max}(\mathbf{H})$. Linear stability requires this principal
curvature to remain below the theoretical limit of $2/\eta$. If the
local curvature exceeds this limit, the optimization step overshoots,
leading to the divergence or violent oscillatory behavior
characteristic of the EoS~\cite{Cohen2021, Damian2023}.

\begin{figure}[t]
\begin{center}
\includegraphics[width=\linewidth]{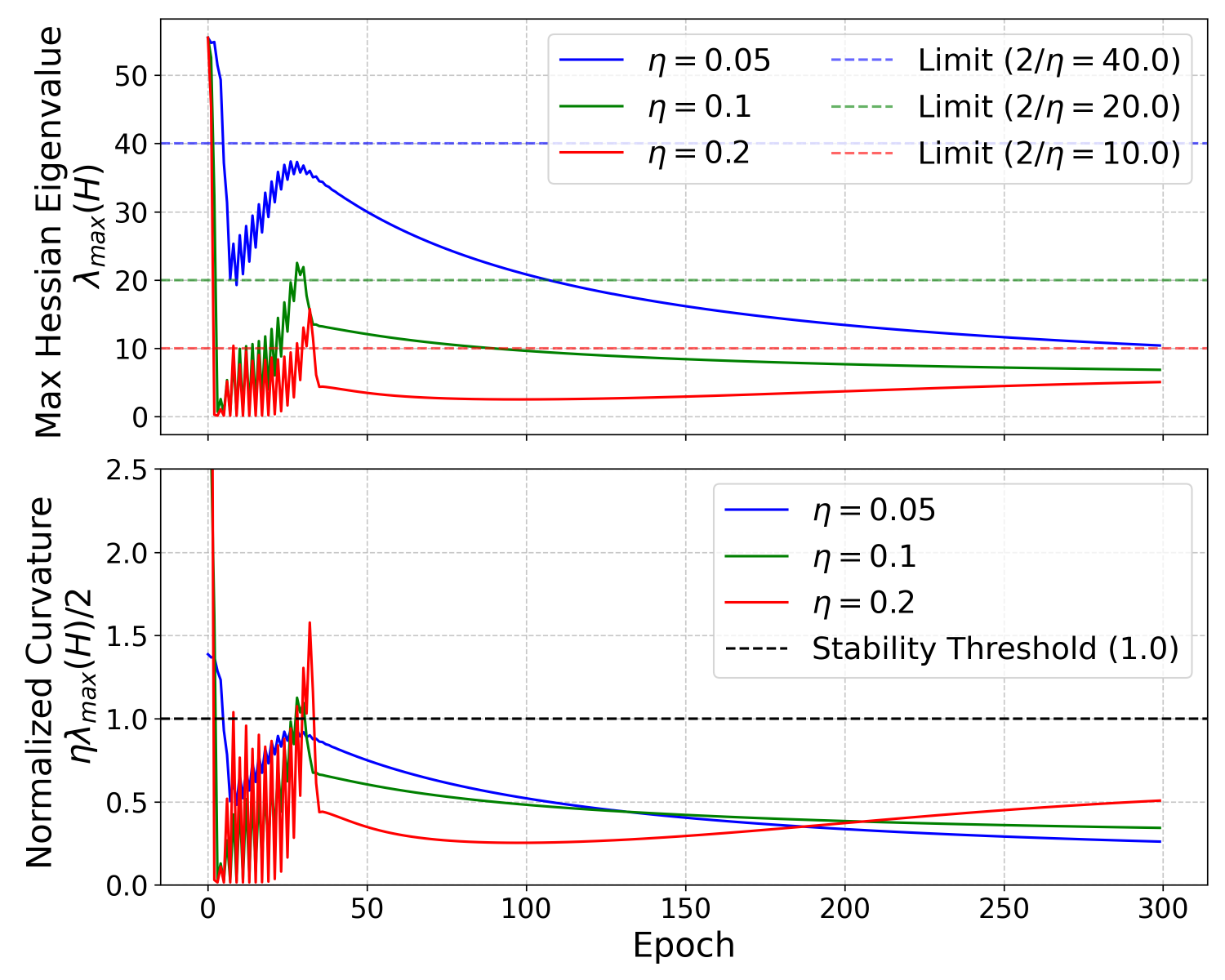}
\caption{Transient self-stabilization of Gradient Descent on the Ridge
  ($N=50$, $P/N=2.0$, $\gamma=0.02$) across different learning rates
  $\eta \in \{0.05, 0.1, 0.2\}$. The trajectories represent a single
  representative trial to clearly illustrate the deterministic
  oscillatory dynamics. \textbf{(Top)} Maximum eigenvalue of the
  Hessian $\mathbf{H}$. The dashed lines indicate the respective
  theoretical stability limits ($2/\eta$). \textbf{(Bottom)}
  Normalized curvature $\eta \lambda_{\max}(\mathbf{H}) / 2$. The
  dashed black line indicates the stability threshold (1.0).}
\label{fig:learning_dynamics}
\end{center}
\end{figure}

To empirically test the relationship between the optimization dynamics
and this stability limit, we tracked the learning process on the Ridge
($P/N=2.0$, $\gamma=0.02$) using three different learning rates:
$\eta \in \{0.05, 0.1, 0.2\}$. For each epoch, we computed the exact
Hessian $\mathbf{H}$ (Eq.~(\ref{eq:hessian})) and extracted its
maximum eigenvalue. Figure~\ref{fig:learning_dynamics} illustrates the
temporal evolution of both the absolute maximum curvature and a
normalized curvature metric, defined as
$\eta \lambda_{\max}(\mathbf{H}) / 2$. This normalized metric
explicitly evaluates how close the system is to the stability limit; a
value equal to or greater than $1.0$ indicates that the system is
operating at or beyond the edge of linear stability.

The learning dynamics exhibit a consistent, three-phase structural
evolution across all tested learning rates:
\begin{enumerate}
\item \textbf{Initial Overshoot and Instability:} At the onset of
  training, the parameters are initialized at zero, where the
  prediction variance is maximized. This results in a massive initial
  curvature ($\lambda_{\max} > 50$) that significantly exceeds the
  stability limit $2/\eta$ for all tested learning
  rates. Consequently, the initial GD update results in a large step
  that overshoots the minimum along the direction of steepest
  curvature, causing the normalized curvature to spike above 1.0
  (bottom panel).
\item \textbf{Transient Self-Stabilization:} Following the overshoot,
  the curvature $\lambda_{\max}$ drops rapidly but is then bounded by
  the stability limit. The top panel clearly shows that the curvature
  for each $\eta$ oscillates precisely just below its corresponding
  limit ($40, 20$, and $10$, respectively). The bottom panel confirms
  this alignment, as the normalized curvature curves for all learning
  rates repeatedly strike the 1.0 threshold before bouncing back. This
  behavior indicates that the system undergoes a transient phase of
  self-stabilization~\cite{Damian2023}, where divergent steps in the
  steep direction dynamically reduce the local curvature, preventing
  the optimization from completely diverging.
\item \textbf{Escape and Stable Convergence:} Importantly, the system
  does not remain permanently trapped at the stability
  boundary. Between epochs 30 and 40, the macroscopic oscillations
  cease. The bottom panel shows that the normalized curvature smoothly
  detaches from the 1.0 threshold and descends into the strictly
  stable regime ($< 1.0$). At this point, the principal curvature has
  naturally decreased sufficiently so that the GD updates are stable
  across all directions, allowing for smooth, monotonic convergence to
  the final solution.
\end{enumerate}

These empirical results demonstrate a profound interaction between the
learning algorithm and the loss landscape. The network parameters do
not passively settle into an existing flat minimum. Instead, when
forced to navigate the extreme curvature near the Ridge, the GD
trajectory relies on a transient self-stabilizing feedback mechanism
to survive the initial instability, eventually carving a path into a
stable, high-capacity representation.

Additional experiments with varying storage loads showed that while
this qualitative three-phase structure remains consistent, the
duration of the transient self-stabilization phase extends
significantly as the load increases (see Appendix A).

\section{Theoretical Analysis of Spectral Collapse and Dynamics}
The empirical observations in the previous sections present two
distinct but related phenomena: the static geometric collapse to a
rank-1 spectral structure at low $\gamma$ values, and the dynamic,
transient self-stabilization near the $2/\eta$ limit during
learning. In this section, we provide analytical interpretations for
these behaviors.

\subsection{Geometric Origin of the Rank-1 Collapse}
The phase diagrams (Fig.~\ref{fig:phase_diagram}) demonstrated that
the statistical manifold loses its multi-dimensional structure as the
kernel locality parameter $\gamma$ decreases, eventually collapsing to
a rank-1 matrix. This asymptotic behavior can be analytically derived
from the Taylor expansion of the RBF kernel.

The RBF kernel between two binary patterns
$\boldsymbol{\xi}^\mu, \boldsymbol{\xi}^\nu \in \{-1, 1\}^N$ depends
on their squared Euclidean distance,
$\|\boldsymbol{\xi}^\mu -~\boldsymbol{\xi}^\nu\|^2 = 2N -
2(\boldsymbol{\xi}^\mu \cdot \boldsymbol{\xi}^\nu)$. Thus, the kernel
matrix elements are:
\begin{equation}
K_{\mu\nu} = \exp\left( -2\gamma N + 2\gamma (\boldsymbol{\xi}^\mu \cdot \boldsymbol{\xi}^\nu) \right).
\end{equation}
In the global regime where $\gamma \ll 1/N$, the first-order Taylor
expansion yields:
\begin{equation}
K_{\mu\nu} \approx 1 - 2\gamma N + 2\gamma (\boldsymbol{\xi}^\mu \cdot \boldsymbol{\xi}^\nu).
\label{eq:kernel_taylor}
\end{equation}
Let $\mathbf{J} = \mathbf{1}\mathbf{1}^\top$ be a $P \times P$ matrix
of ones, and $\mathbf{X} \in \{-1, 1\}^{P \times N}$ be the pattern
matrix. Equation (\ref{eq:kernel_taylor}) can be written in matrix
form as:
\begin{equation}
\mathbf{K} \approx (1 - 2\gamma N)\mathbf{J} + 2\gamma \mathbf{X} \mathbf{X}^\top.
\label{eq:kernel_matrix_approx}
\end{equation}

The FIM is defined as $\mathbf{G} = \mathbf{K} \mathbf{D}
\mathbf{K}$. In the limit $\gamma \to 0$, the kernel matrix is
dominated by the constant matrix, $\mathbf{K} \to
\mathbf{J}$. Substituting this into the FIM expression yields:
\begin{equation}
  \mathbf{G} \approx \mathbf{J} \mathbf{D} \mathbf{J} = (\mathbf{1}\mathbf{1}^\top) \mathbf{D} (\mathbf{1}\mathbf{1}^\top) = \mathrm{Tr}(\mathbf{D}) \mathbf{J}.
\label{eq:fim_limit}
\end{equation}
The matrix $\mathbf{J}$ has a rank of 1, with a single non-zero
eigenvalue of $P$ and the corresponding eigenvector
$\mathbf{1}$. Therefore, as $\gamma \to 0$, the maximum eigenvalue of
the FIM approaches
$\lambda_{\max}(\mathbf{G}) \approx P \cdot \mathrm{Tr}(\mathbf{D})$,
while all other eigenvalues approach zero. Since the regularization
$\lambda$ is small, the Hessian $\mathbf{H}$ exhibits a similar
extreme spectral degeneracy.

This derivation clarifies why the functional high-capacity regime (the
Ridge) is located just before this collapse. The first term in
Eq.~(\ref{eq:kernel_matrix_approx}) acts as a uniform restorative
force (generating the massive $\lambda_{\max}$), while the second term,
corresponding to classical Hebbian interactions
$\mathbf{X}\mathbf{X}^\top$, preserves the dimensionality required for
pattern discrimination. Memory performance is optimal when these two
opposing structural forces are balanced at the critical boundary.

\subsection{Qualitative Model of Self-Stabilization}
Next, we provide a theoretical interpretation of the transient
self-stabilization observed in Fig.~\ref{fig:learning_dynamics}. As
noted in recent literature~\cite{Damian2023}, the EoS behavior can be
understood through higher-order interactions where divergence along a
steep direction causes a reduction in local curvature.

In the KLR setting, this self-stabilizing negative feedback loop is
explicitly mediated by the properties of the logistic variance. Let
$x$ represent the magnitude of the parameter component along the
principal eigenvector $\mathbf{v}_{\max}$ of the Hessian. As a
qualitative reduced-order model, the GD update along this specific
dominant direction can be approximated by a linearized step scaled by
the local principal curvature $\lambda_{\max}(x)$:
\begin{equation}
x_{t+1} \approx x_t - \eta \lambda_{\max}(x_t) x_t = (1 - \eta \lambda_{\max}(x_t)) x_t.
\label{eq:gd_step}
\end{equation}

A fundamental property of the KLR objective is that the curvature
$\lambda_{\max}$ is proportional to the trace of the prediction
variance matrix $\mathbf{D}$ (as seen in
Eq.~(\ref{eq:fim_limit})). The variance for each pattern,
$p_\mu(1-p_\mu)$, is maximized when the prediction is completely
uncertain ($p_\mu=0.5$, which occurs when parameters are near
zero). Consequently, as the magnitude of the parameters $|x|$
increases and the model's predictions become more confident
($p_\mu \to 1$ or $p_\mu \to 0$), the overall variance
decreases. Therefore, the principal curvature $\lambda_{\max}(x)$ acts
as a monotonically decreasing function of $|x|$.

This non-linear relationship creates a self-stabilizing dynamic feedback loop:
\begin{enumerate}
\item \textbf{Instability and Divergence:} When initialized near zero,
  $\lambda_{\max}(0) > 2/\eta$. The multiplier
  $(1 - \eta \lambda_{\max}(x_t))$ in Eq.~(\ref{eq:gd_step}) becomes
  less than $-1$, causing the parameter $x_t$ to flip its sign and its
  magnitude $|x_t|$ to grow exponentially.
\item \textbf{Curvature Suppression (Self-Stabilization):} As the
  magnitude $|x_t|$ increases due to this divergence, the logistic
  variance decreases, which in turn reduces the local curvature
  $\lambda_{\max}(x_t)$.
\item \textbf{Boundary Equilibration:} The magnitude $|x_t|$ continues
  to grow until the curvature is suppressed exactly to the stability
  boundary, $\lambda_{\max}(x_t) \approx 2/\eta$. At this boundary, the
  multiplier approaches $-1$, resulting in stable oscillatory behavior
  ($x_{t+1} \approx -x_t$).
\end{enumerate}

While Eq.~(\ref{eq:gd_step}) is a simplified one-dimensional model, it
effectively captures the essence of the mechanism. The initial
overshoot forces the network parameters away from the high-variance
origin. This divergence inherently suppresses the maximum curvature
until it respects the learning rate limit, allowing the optimization
process to survive the singular geometry of the Ridge and eventually
escape into a stable region as the global loss decreases.

\section{Discussion}
Our geometric and dynamic analyses provide a physical interpretation
of how high-capacity kernel associative memories are formed. In this
section, we discuss the implications of these findings for learning
theory, relate them to recent studies on optimization dynamics, and
outline the limitations of our current approach.

\subsection{Rethinking the Flat Minima Hypothesis}
A prominent hypothesis in deep learning optimization suggests that
optimization algorithms preferentially converge to ``flat minima,''
which are characterized by low curvature (small eigenvalues of the
Hessian or FIM) and are generally associated with better
generalization capabilities~\cite{Hochreiter1997,
  Keskar2017}. However, our findings in KLR-trained Hopfield networks
present a distinct scenario for this established view.

As shown in the phase diagram (Fig.~\ref{fig:phase_diagram}), the
optimal operating regime for associative memory, the Ridge of
Optimization, is not located in a flat region of the parameter
space. Instead, it is situated adjacent to a region of extreme
spectral degeneracy, characterized by a sharp curvature in the
dominant principal direction. Learning successfully converges
precisely when the trajectory navigates this highly curved
region. This suggests that for structured representation tasks like
associative memory, optimal performance may not always correlate with
minimizing curvature across all dimensions. Rather, the systematic
amplification of a specific principal curvature can be beneficial for
creating a deep, globally attractive energy basin, provided a minimal
set of non-zero trailing eigenvalues is preserved for pattern
discrimination.

\subsection{The Margin-Seeking Tendency and Spectral Concentration}
The drive toward the highly curved Ridge geometry can be conceptually
related to the implicit bias of the optimization objective. It is
well-established that optimizing unregularized logistic-type loss
functions on separable data leads to parameters that asymptotically
maximize the classification margin~\cite{Rosset2004, Soudry2018}.

While our objective function (Eq.~(\ref{eq:objective})) includes an
explicit $L_2$ regularization term, the margin-seeking tendency of the
logistic component provides a plausible explanation for the observed
spectral concentration. In the context of kernel Hopfield networks,
creating robust attractors requires separating each stored pattern
from all others in the high-dimensional feature space. As the learning
algorithm minimizes the logistic loss, it pushes the weight vectors
toward larger magnitudes to increase confidence. This behavior
actively stretches the parameter space along the principal directions
responsible for class separation. The emergence of the near-singular
Ridge geometry and the resulting extreme disparity in the Hessian
eigenvalue spectrum can thus be understood as a natural consequence of
gradient descent operating on margin-based loss functions, balanced by
the regularizer.

\subsection{Transient Self-Stabilization and the Edge of Stability}
Our dynamic analysis in Section 5.2 explains how GD interacts with the
stability limit ($2/\eta$) using a discrete-time difference equation
derived from the properties of the logistic variance. This phenomenon
is closely related to the general theory of ``self-stabilization''
recently proposed by Damian et al.~\cite{Damian2023}. Using a
third-order Taylor expansion, they demonstrated that the EoS can be
modeled as a negative feedback dynamical system, where divergence
along the dominant eigenvector causes the local curvature to decrease
until stability is temporarily restored.

Our findings are consistent with this theoretical framework. While
Damian et al. provide a general mechanism based on higher-order
derivatives, our analysis offers a concrete, domain-specific
realization of this self-stabilization loop. In the KLR model, the
negative feedback is explicitly mediated by the prediction variance
$p(1-p)$ inherent to the FIM and Hessian: a gradient overshoot
increases the parameter magnitude, which immediately suppresses the
probability variance, thereby reducing the local
curvature. Recognizing this connection suggests that the transient EoS
phase observed in our associative memory model is a fundamental,
self-correcting property of gradient descent navigating sharp
landscapes.

\subsection{Implications for Curvature-Aware Optimization}
The oscillatory dynamics of GD (Fig.~\ref{fig:learning_dynamics})
highlight the fundamental challenge of navigating the highly
anisotropic landscape near the Ridge. GD, which operates based on the
Euclidean metric, is forced into a transient oscillatory phase because
the principal curvature initially exceeds its stability limit.

This behavior provides an information-geometric motivation for
utilizing curvature-aware optimization methods, such as Natural
Gradient Descent (NGD)~\cite{Amari2016}. By preconditioning the update
direction with the inverse of the FIM~($\mathbf{G}^{-1}$), NGD can
theoretically neutralize extreme disparities in curvature. However, as
our static analysis shows, the Ridge is located near a region of
spectral degeneracy where the FIM approaches a rank-deficient
state. In such regimes, directly applying the inverse FIM is
computationally unstable.

Consequently, practical implementations of natural-gradient-type
methods for high-capacity memory require robust damping mechanisms or
pseudo-inverse treatments to safely navigate the singularity. Indeed,
recent work has empirically demonstrated that NGD, when appropriately
regularized with a damping factor, can successfully bypass the
oscillatory EoS phase and follow a smooth, geodesic-like trajectory on
the Ridge~\cite{tamamori_nolta_ngd}. The present analysis complements
those findings by elucidating the underlying cause of the GD
instability, specifically the transient self-stabilization triggered
by the interaction between the extreme principal curvature and the
learning rate limit.

\subsection{Limitations and Future Work}
While our analysis clarifies the geometric self-organization
mechanism, several limitations must be acknowledged:
\begin{enumerate}
\item \textbf{Restriction to Uncorrelated Patterns:} The present study
  focuses on networks storing independent random binary patterns. This
  idealized setting isolates the fundamental spectral behaviors, such
  as the asymptotic rank-1 collapse represented by matrix
  $\mathbf{J}$. However, real-world data typically exhibit structured
  correlations. These correlations are likely to alter the structure
  of the spectral degeneracy, potentially forming hierarchical,
  low-rank spectra rather than a strict rank-1 collapse.
\item \textbf{Qualitative Nature of the Dynamic Model:} The
  theoretical model presented in Section 5.2 uses a simplified
  one-dimensional difference equation to explain the interaction with
  the $2/\eta$ limit. While this provides a qualitative understanding
  of the feedback mechanism, a rigorous, high-dimensional dynamical
  systems analysis is required to precisely characterize the
  interaction between the dominant EoS oscillations and the
  convergence along the trailing eigenvectors.
\end{enumerate}
Based on these limitations, extending the spectral analysis to
datasets with semantic correlations and developing scalable,
curvature-aware optimization algorithms that can robustly handle
spectral degeneracy are promising directions for future research.

\section{Conclusion}
In this study, we investigated the geometric and dynamical mechanisms
underlying memory formation in high-capacity KLR Hopfield networks. By
integrating static spectral analysis with an examination of learning
trajectories, we provided a physical interpretation of the
optimization process near the storage limit.

Our static analysis revealed that the previously identified ``Ridge of
Optimization'' aligns with the boundary of a region characterized by
extreme spectral degeneracy. We analytically demonstrated that as the
kernel mapping becomes increasingly global, the Fisher Information
Matrix asymptotically approaches a rank-1 structure. The functional
high-capacity regime is situated adjacent to this geometric
transition, balancing the massive principal curvature required for
global stability with the spectral diversity needed for pattern
separation.

Furthermore, we demonstrated that the learning trajectory of GD on the
Ridge is fundamentally shaped by the EoS phenomenon. Rather than
converging monotonically to a flat minimum, the network parameters
undergo a phase of transient self-stabilization. As predicted by a
qualitative reduced-order model, the negative feedback inherent to the
logistic prediction variance forces the maximum eigenvalue of the
Hessian to dynamically equilibrate near the stability limit dictated
by the learning rate. This self-correcting mechanism allows the
optimization process to survive the initial instability and carve out
an exceptionally sharp principal attractor basin.

These findings suggest that optimal memory representations in
high-dimensional kernel spaces are dynamically formed at the highly
curved boundaries of geometric singularities. This understanding not
only clarifies the resilience of standard gradient methods in
pathological landscapes but also reinforces the motivation for
exploring curvature-aware optimization algorithms capable of
efficiently navigating these degenerate spaces.

\appendix
\section{Consistency of Self-Stabilization Across Storage Loads}
In Section 4, we analyzed the learning dynamics of GD and identified a
transient self-stabilization mechanism near the stability limit
$2/\eta$. This analysis was conducted at a representative storage load
of $P/N = 2.0$. To verify that this dynamical behavior is a consistent
feature of the optimization process on the Ridge, rather than an
artifact of a specific load, we evaluated the learning trajectories
across higher storage capacities.

\begin{figure}[t]
\begin{center}
\includegraphics[width=\linewidth]{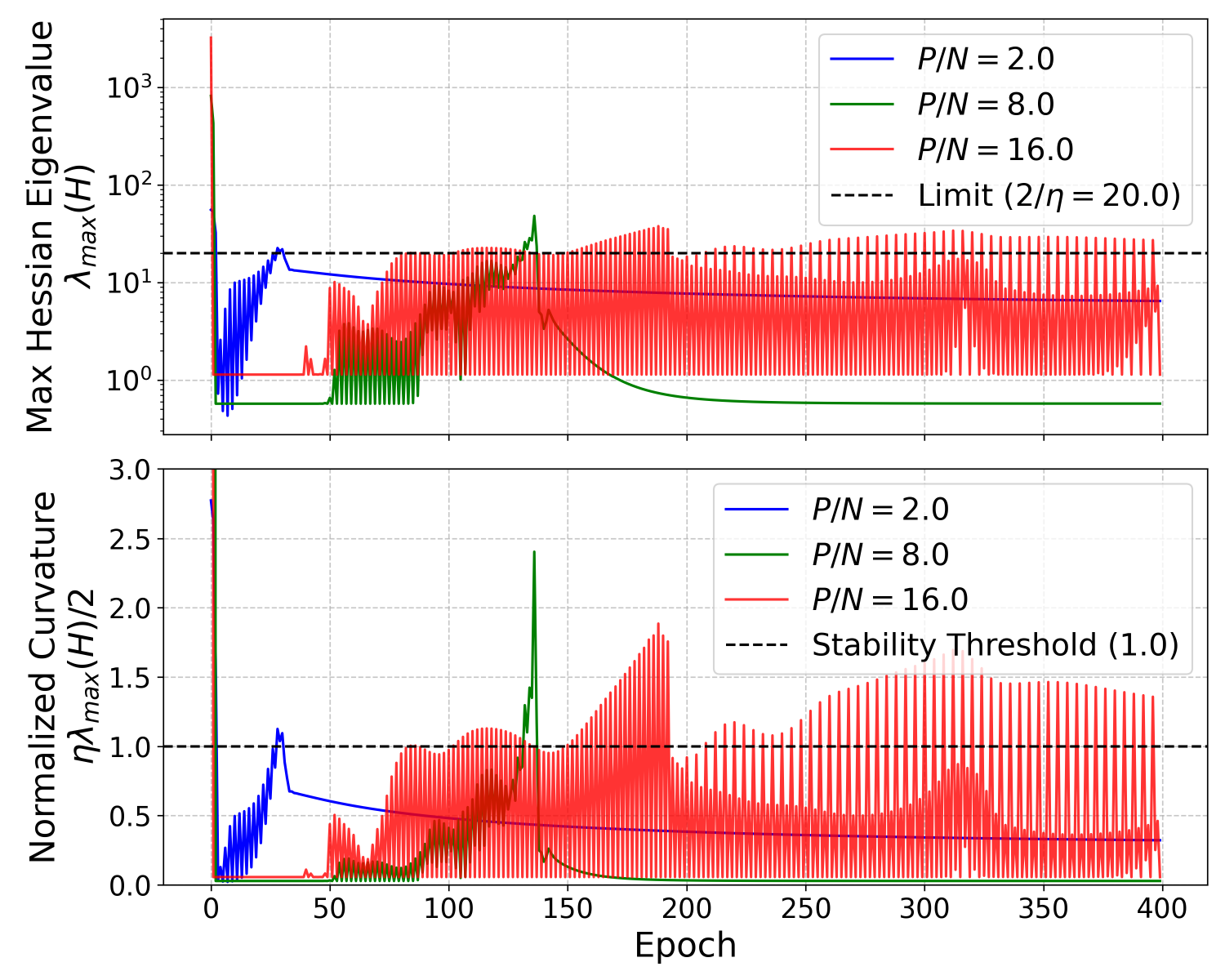}
\caption{Learning dynamics of GD on the Ridge across varying storage
  loads ($P/N \in \{2.0, 8.0, 16.0\}$) at a fixed learning rate
  ($\eta=0.1$). \textbf{(Top)} Maximum Hessian eigenvalue
  $\lambda_{\max}(\mathbf{H})$ on a logarithmic scale. The initial
  curvature spike increases proportionally with the storage load
  $P$. \textbf{(Bottom)} Normalized curvature
  $\eta \lambda_{\max}(\mathbf{H}) / 2$. Regardless of the load, the
  trajectory undergoes a transient self-stabilization phase near the
  1.0 threshold before eventually escaping into the stable convergence
  regime.}
\label{fig:appendix_dynamics}
\end{center}
\end{figure}

Figure \ref{fig:appendix_dynamics} presents the evolution of the
maximum Hessian eigenvalue $\lambda_{\max}(\mathbf{H})$ and the
normalized curvature $\eta \lambda_{\max}(\mathbf{H}) / 2$ for networks
storing $P/N \in \{2.0, 8.0, 16.0\}$, keeping the kernel locality
fixed at $\gamma = 0.02$ and the learning rate at $\eta = 0.1$.

The results demonstrate the consistency of the three-phase learning
structure across varying degrees of memory congestion, while
highlighting two key load-dependent characteristics:
\begin{enumerate}
\item \textbf{Scaling of the Initial Curvature:} As analytically
  derived in Section 5.1 (Eq.~(\ref{eq:fim_limit})), the maximum
  eigenvalue of the FIM near the origin is proportional to the number
  of stored patterns $P$. The top panel of
  Fig.~\ref{fig:appendix_dynamics} empirically confirms this
  relationship. At higher loads (e.g., $P/N = 16.0$), the initial
  spike in $\lambda_{\max}(\mathbf{H})$ reaches orders of magnitude
  higher than at $P/N = 2.0$.
\item \textbf{Prolongation of the Oscillatory Phase:} Because the
  initial curvature is substantially larger at higher loads, the
  self-stabilization feedback loop (which suppresses the curvature by
  driving parameters away from zero) requires more iterations to bring
  the principal curvature down to the stability limit. The bottom
  panel shows that while the $P/N = 2.0$ trajectory escapes the 1.0
  threshold rapidly, the $P/N = 16.0$ trajectory remains in the
  oscillatory search phase for a significantly longer duration before
  achieving stable convergence.
\end{enumerate}

These supplementary observations validate our theoretical model. The
EoS is a robust and necessary phase of learning in high-capacity
kernel associative memories. The optimization algorithm must undergo
this transient self-stabilization to navigate the extreme,
load-dependent curvature of the singularity before it can successfully
form deep and stable attractor basins.

\profile{Akira Tamamori}{He received his B.E., M.E., and D.E. degrees
  from Nagoya Institute of Technology, Nagoya, Japan, in 2008, 2010,
  2014, respectively.  From 2014 to 2016, he was a Research Assistant
  Professor at Institute of Statistical Mathematics, Tokyo,
  Japan. From 2016 to 2018, he was a Designated Assistant Professor at
  the Institute of Innovation for Future Society, Nagoya University,
  Japan. From 2018 to 2020, he was a Junior Associate Professor at
  Aichi Institute of Technology, Japan.  He has been an Associate
  Professor at Aichi Institute of Technology since 2020. He is a
  member of the Institute of Electronics, Information and
  Communication Engineers (IEICE), the Information Processing Society
  of Japan (IPSJ), the Acoustical Society of Japan (ASJ), and
  Asia-Pacific Signal and Information Processing Association
  (APSIPA).}

\end{document}